\documentclass[11pt]{article}
\usepackage[final]{acl}
\usepackage{times}
\usepackage{latexsym}
\usepackage[T1]{fontenc}
\usepackage[utf8]{inputenc}
\usepackage{microtype}
\usepackage{booktabs}
\usepackage{graphicx}
\usepackage{amssymb}
\usepackage{tikz}
\usetikzlibrary{positioning,fit,backgrounds,calc,arrows.meta}

\title{QuanReview: Offline, Auditable Reconciliation of Human and LLM Span Annotations}

\author{
  Matteo Musacchio$^{1}$, \enspace
  Juan Cruz Giner Pulero$^{1}$, \enspace
  Isabel Casta{\~n}eda$^{1}$, \enspace
  Naomi Couriel$^{1}$, \\
  Yelena Mejova$^{2}$, \enspace
  Mariano G. Beir\'o$^{1,4}$, \enspace
  Kyriaki Kalimeri$^{2,3}$
   \\[4pt]
  \normalsize $^{1}$Universidad de San Andr\'es, Buenos Aires, Argentina \\
  \normalsize $^{2}$ISI Foundation, Turin, Italy \\ \qquad
  $^{3}$UNICEF, New York, NY, USA \\
  \qquad
  $^{4}$CONICET, Buenos Aires, Argentina}
\date{}

\begin{document}
\maketitle

\begin{abstract}
Structured span annotations, such as quantities with their units,
uncertainty modifiers, and event classes, are expensive to create and
hard to keep trustworthy once language models enter the loop. We
present QuanReview, an open-source system for auditing and correcting
such annotation layers. QuanReview aligns two annotation streams over
the same documents at character level, resolves unambiguous cases by
an explicit and logged policy, and routes candidate conflicts to a
browser-based adjudication interface where reviewers accept either
side, build field-level hybrids, or flag items for re-annotation. A
campaign manager assigns documents to multiple annotators with
configurable redundancy, computes agreement at document and span
level, auto-merges unanimous documents, and exports the corrected
layer in the original file format, so that it can replace the
original annotation files directly. Applied to a
4{,}457-record humanitarian benchmark and an LLM extraction stream,
the system fully auto-merged 8\% of documents, applied automatic policy decisions to a
further 1{,}513 records, and concentrated human attention on 3{,}131
candidate conflicts, a mean of 5.4 per reviewed document.
\end{abstract}

\section{Introduction}

Dataset curators eventually discover that their reference annotations
disagree with what a strong language model extracts from the same
text. Some disagreements are model errors, some are reference errors,
and some are genuine ambiguities. Deciding which is which, at scale
and without corrupting the dataset, is a workflow problem rather than
a modelling problem. As language models are increasingly used both
to produce and to consume span-level annotation layers, keeping those
layers trustworthy has become a first-class concern for the NLP
community: benchmark leakage, silver-labelled training data, and
LLM-assisted re-annotation all depend on being able to reconcile
human and model annotations with an auditable trail. Every record
needs a position-anchored comparison between the two versions,
one-sided cases need an explicit policy instead of silent
handling,
real conflicts need a human decision, and all of it needs an audit
trail so the corrected dataset can be trusted later.

QuanReview packages this workflow into one system. It was built to
audit the quantitative annotation layer of
HumSet~\citep{fekih2022humset, liberatore2024quantitative}, where
each record binds a number to its unit, uncertainty modifier, and
event class, but the schema is configurable and the workflow applies
to other typed-span annotation tasks. Its target users are dataset
curators who maintain span-annotation layers, NLP researchers who
need defensible corrected references, and domain experts, in our case
humanitarian analysts, who review model output through a browser
without touching JSON.\footnote{Code and a runnable demo:
\url{https://github.com/mattemusacchio/quanreview}. 
}

\section{Related systems}

General-purpose annotation platforms support parts of this problem.
INCEpTION~\citep{klie2018inception} includes a curation stage that
compares and merges multiple annotators' work, and Label
Studio~\citep{tkachenko2020labelstudio} documents workflows for
inter-annotator agreement and for comparing model predictions with
annotations; doccano~\citep{nakayama2018doccano} and
Prodigy~\citep{montani2018prodigy} focus on annotation from raw text
and model-assisted labelling, while Argilla~\citep{argilla2023}
targets LLM-in-the-loop data curation. Presenting automated
recommendations alongside human judgements can introduce automation
and anchoring biases~\citep{skitka1999automation}, motivating our use
of neutral stream labels, randomised presentation order, reversible
decisions, and explicit conflict routing.
QuanReview does not replace these platforms and was itself preceded
by INCEpTION~\citep{klie2018inception} in creating the reference
layer. Its contribution is a specialised, lightweight combination
for post-hoc reconciliation of two complete annotation exports:
local, file-first offline operation with no external server or hosted service,
deterministic character-offset alignment of finished streams,
explicit and configurable policies for one-sided records,
conflict-only routing instead of re-annotation, field-level hybrid
decisions, reversible record-level audit logs, and export in the
input schema and file layout so that the corrected layer replaces the
original files directly.
Table~\ref{tab:related-systems} focuses on systems that support some
form of adjudication or multi-annotator workflow; from-scratch
annotation tools such as brat~\citep{stenetorp2012brat} and doccano
are omitted from the table because they have no stage for reconciling
two existing annotation sets. We have not yet compared QuanReview
with an established platform on the same curation task; such a
controlled head-to-head comparison is left for future work.

\begin{table*}[!htbp]
  \centering
  \small
  \setlength{\tabcolsep}{5pt}
  \begin{tabular}{lccccc}
  \toprule
  \textbf{System} & \textbf{Two-stream} & \textbf{Field-level} & \textbf{Char-level} & \textbf{IAA /} &
  \textbf{Decision} \\
                  & \textbf{reconcile}  & \textbf{hybrid}      & \textbf{align}      & \textbf{campaign} &
  \textbf{audit log} \\
  \midrule
  Label Studio    & partial    & --         & --         & partial    & partial    \\
  INCEpTION       & partial    & partial    & --         & \checkmark & partial    \\
  Prodigy         & partial    & --         & --         & --         & partial    \\
  Argilla         & partial    & --         & --         & partial    & \checkmark \\
  \midrule
  \textbf{QuanReview} & \checkmark & \checkmark & \checkmark & \checkmark & \checkmark \\
  \bottomrule
  \end{tabular}
  \caption{Feature comparison across annotation and adjudication tools. QuanReview is the only system combining
  two-stream reconciliation, field-level hybrids, and character-level alignment. \textit{Partial} denotes support for a related capability but not the complete post-hoc two-stream workflow defined here.}
  \label{tab:related-systems}
  \end{table*}

\section{System overview}

\begin{figure*}[t]
\centering
\begin{tikzpicture}[
  font=\scriptsize,
  >={Stealth[length=1.6mm,width=1.3mm]},
  box/.style={draw, line width=0.5pt, align=center, inner sep=3pt,
              minimum height=7mm, text width=22mm, fill=white},
  arr/.style={->, line width=0.5pt},
  dsh/.style={arr, dashed, dash pattern=on 1.6pt off 1.3pt},
  joint/.style={draw, circle, minimum size=1.7mm, inner sep=0pt, fill=white},
  glab/.style={font=\scriptsize\bfseries\itshape, text=green!35!black},
  note/.style={font=\tiny\itshape, text=green!35!black},
]
\node[box, text width=23mm] (gt)  {Reference annotation\\\textcolor{blue!60!black}{HumSet spans}};
\node[box, text width=23mm, below=3mm of gt] (llm) {Model output\\\textcolor{violet!70!black}{LLM spans}};
\coordinate (in) at ($(gt.east)!0.5!(llm.east)$);
\node[joint, right=5mm of in] (j) {};
\draw[line width=0.5pt] (gt.east)  -| (j.north);
\draw[line width=0.5pt] (llm.east) -| (j.south);

\node[box, right=5mm of j, text width=19mm] (align) {Alignment core\\[1pt]{\tiny\itshape character-level pairing}};
\draw[arr] (j) -- (align);

\node[box, right=9mm of align, yshift=8mm,  text width=27mm] (auto)
  {Policy engine\\[1pt]{\tiny\itshape reference-only keep / model-only defer}};
\node[box, right=9mm of align, yshift=-8mm, text width=27mm] (rev)
  {Adjudication UI\\[1pt]{\tiny\itshape reference vs.\ model conflicts}};
\draw[arr] (align.east) -- ++(4mm,0) |- (auto.west);
\draw[arr] (align.east) -- ++(4mm,0) |- (rev.west);

\begin{scope}[on background layer]
  \node[draw=green!45!black, line width=0.5pt, fill=green!8,
        inner sep=3.5mm, fit=(auto)(rev)] (grp) {};
\end{scope}
\node[glab, above=0.3mm of grp.north] {Conflict resolution};

\coordinate (m) at ($(auto.east)!0.5!(rev.east)$);
\node[box, right=7mm of m, text width=23mm] (merge) {Campaign merge\\[1pt]{\tiny\itshape auto-merge unanimous}};
\node[box, right=7mm of merge, text width=18mm, fill=green!14,
      draw=green!45!black] (out) {Corrected\\layer};
\draw[arr] (auto.east) -| ($(m)+(5mm,0)$) -- (merge.west);
\draw[arr] (rev.east)  -| ($(m)+(5mm,0)$) -- (merge.west);
\draw[arr] (merge) -- (out);

\node[box, below=8mm of rev, text width=31mm] (flag)
  {Flagged: unaligned pair,\\possible new sample};
\draw[arr] (rev.south) -- (flag.north);
\draw[dsh] (flag.west) -| (align.south)
  node[note, pos=0.65, right=1mm, anchor=west] {Second annotation round};
\end{tikzpicture}
\caption{System architecture. Reference (HumSet) and model (LLM)
spans are aligned at character level; the policy engine auto-resolves
one-sided records (reference-only kept, model-only deferred) and
routes conflicting pairs to the adjudication UI. Reviewed documents
are merged across annotators, unanimous ones automatically, into the
corrected layer, which replaces the original files directly; flagged pairs
and deferred model-only records are queued for a second annotation
round.}
\label{fig:pipeline}
\end{figure*}
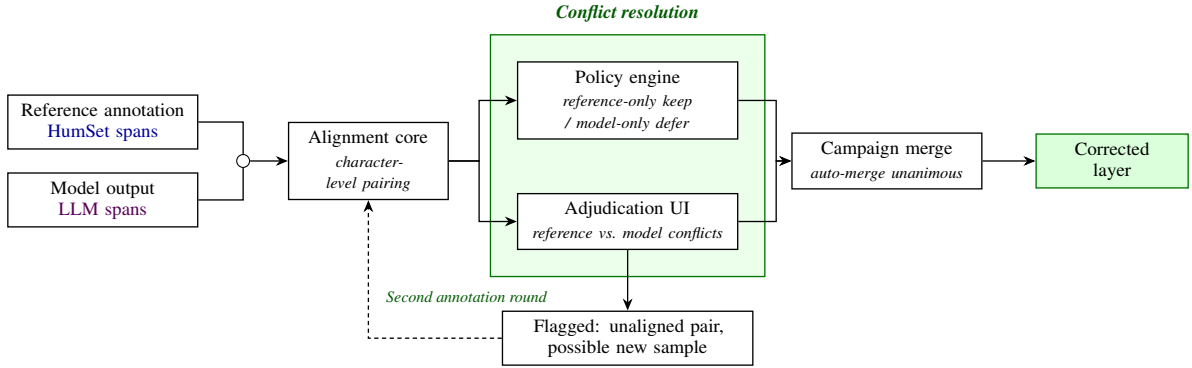

QuanReview is a Python application with a browser interface (Eel);
it runs fully offline on local files, and no document text or
annotation leaves the machine. It has five components, wired into the
correction pipeline of Figure~\ref{fig:pipeline}.

\paragraph{Alignment core.} A deterministic engine pairs the two
streams per document on overlapping character offsets and compares
matched records field by field. The fields and their comparison paths
are declared in a YAML schema file; the quantity schema used in our
campaigns (quantity, unit, modifier, event type) is one instance; the
schema is declarative, allowing additional typed-span and categorical
annotation schemas to be specified through the same configuration
mechanism.

\paragraph{Policy engine.} One-sided records are resolved by explicit,
logged policy rather than silently. Reference-only records are kept.
Model-only records are not discarded: because a share of them may be
genuine events the reference layer missed rather than model
hallucinations, they are automatically deferred and queued for a
second annotation round rather than added to the corrected layer. Every
automatic policy decision is logged, and every reviewer decision is
reversible.

\begin{figure}[t]
  \centering
  \includegraphics[width=\columnwidth]{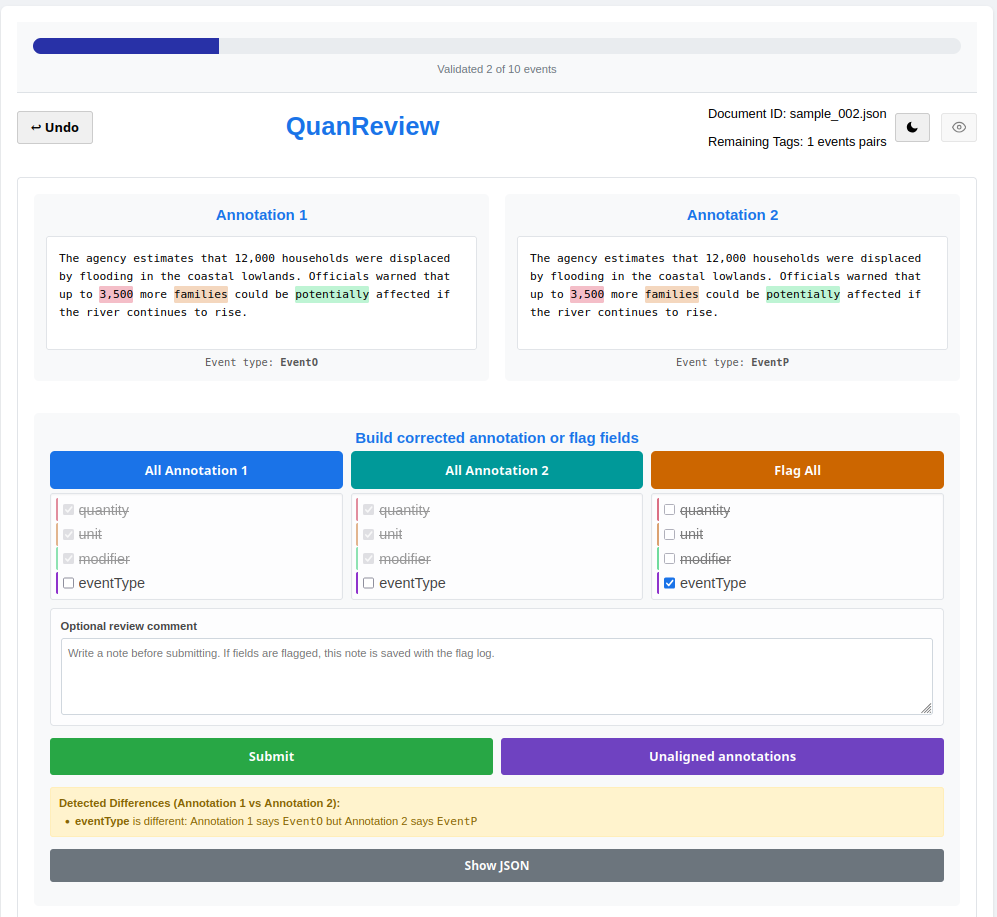}
  \caption{The adjudication interface, showing one conflict pair. The
  two candidates appear under neutral, order-randomised labels; in
  this pair they agree on quantity, unit, and modifier but differ on
  event type (\texttt{EventO} vs.\ \texttt{EventP}), the disagreement
  the reviewer resolves below.}
  \label{fig:adjudication}
\end{figure}

\paragraph{Adjudication interface.} Candidate conflicts appear one at
a time in a two-column browser view showing the source text with both
spans highlighted (Figure~\ref{fig:adjudication}). The reviewer
accepts either side, constructs a field-by-field hybrid, or flags the
pair. A flag is an explicit ``neither candidate is acceptable and I
cannot resolve this from the two streams alone'' signal: the pair is
withheld from the corrected layer and queued for a fresh second
annotation round on the original text, rather than the reviewer being
forced to guess. Flagging is thus a routing action that sends hard
cases back to annotation; it is not itself a correction. Undo,
per-document review state, and free-text review comments are built in,
and every decision is written to disk immediately.

\paragraph{Campaign manager.} A configuration file declares the
annotators, the redundancy level, and an optional validation split.
The orchestrator assigns only documents that actually contain
discrepancies, balances load across annotators, and gives each of
them a one-command launcher that shows only their assignment. The
integration step computes Fleiss'
$\kappa$~\citep{fleiss1971kappa} over adjudication decisions,
auto-merges documents on which all assigned annotators agree, and
emits a conflict bundle per disagreement that a resolution
command-line tool closes out.

\paragraph{Export layer.} Corrected documents are written in the same
JSON schema as the input, together with per-document review-state
files and comment logs. The corrected layer can therefore replace
the original annotation files directly, with provenance for every
changed record.

\section{User workflows}

\paragraph{Single-reviewer audit.} Point the tool at a reference
directory and a model-output directory, review the conflict queue,
export the corrected layer. This is the fastest way to answer the
question ``how much should I trust this annotation layer?''

\paragraph{Multi-annotator campaign.} Declare annotators and
redundancy in the configuration file, run the orchestrator, and let
each annotator launch their own session. Integration reports
agreement, merges unanimous documents, and routes the remainder to an
adjudicator.

\paragraph{Model audit.} Load two model outputs instead of a
reference and a model to inspect where systems disagree before
spending any human annotation time. The same alignment, policies, and
interface apply.

\section{Demonstration scenario}

The live demonstration walks through a miniature campaign on
humanitarian situation-report excerpts. We configure the schema,
launch two reviewer sessions, and adjudicate two characteristic
conflicts: a quantity whose event class differs between the streams,
and a record where one stream missed the uncertainty modifier. We
also show a model-only record the reference layer missed being
automatically deferred to a second annotation round rather than
silently dropped. We then run integration, inspect the agreement report, and
open the exported corrected files next to the originals to show that
they share the same schema and file layout and can replace them
directly.
Visitors can drive the interface themselves on the sample dataset.

  \section{System validation}

  We report workflow-level evidence of what the system automates and
  what it asks of people.

  Each annotated record in this benchmark is a typed span described by
  four fields, which are also the fields QuanReview compares stream
  against stream: the \emph{quantity} (the measured value and its text
  span), its \emph{unit} of measurement, an \emph{uncertainty
  modifier} (e.g.\ \emph{about}, \emph{at least}, \emph{up to}), and
  the \emph{event type} that classifies what the quantity refers to.
  The per-field agreement in Table~\ref{tab:fleiss} is reported over
  exactly these four fields.

  Aligning the HumSet quantitative reference layer (4{,}457 records
  over 635 documents) with a GPT-4.1 extraction stream, QuanReview
  fully auto-merged 53 documents (8\%) that contained no conflicting
  pairs, applied automatic policy decisions to a further 1,513 records inside reviewed documents (792 exactly matching records retained, 434 reference-only records retained, and 287 model-only records excluded from the corrected layer and queued for re-annotation), and routed 3{,}131 candidate conflict pairs to reviewers,
  a mean of 5.4 per reviewed document.
  Model-only records are preserved separately because some may correspond to genuine annotations missing from the reference layer.

  Eight reviewers from two institutions adjudicated conflicts through
  the interface, with an overlap subset of five reviewers assigned to
  a common set of ten documents (107 shared annotations) used to
  estimate inter-annotator agreement.
  Table~\ref{tab:annotator-activity} summarises per-reviewer activity
  at annotation level; flag rates range from 2.0\% to 24.5\% of pairs,
  isolating the ambiguous cases that the flag mechanism is meant to
  catch.

\begin{table}[!htbp]
  \centering
  \small
  \setlength{\tabcolsep}{4pt}
  \begin{tabular}{lcccccc}
  \toprule
  \textbf{Rev.} & \textbf{Docs} & \textbf{Pairs} & \textbf{Anns} &
  \textbf{Flags} & \textbf{Flag} & \textbf{Changed} \\
                &               &                &               &
                & \textbf{rate} & \textbf{vs ref.} \\
  \midrule
  R1 & 141 & 745 & 996 & 166 & 22.3\% & 42.3\% \\
  R2 & 138 & 761 & 609 &  69 &  9.1\% & 20.9\% \\
  R3 & 133 & 703 & 978 &  14 &  2.0\% & 71.3\% \\
  R4 & 118 & 580 & 539 &  85 & 14.7\% & 12.4\% \\
  R5 &  65 & 352 & 406 &  53 & 15.1\% & 52.8\% \\
  R6 &  34 & 193 & 266 &  36 & 18.7\% & 62.7\% \\
  R7 &  28 & 139 & 194 &  34 & 24.5\% & 59.7\% \\
  R8 &  27 & 109 & 172 &  24 & 22.0\% & 56.9\% \\
  \bottomrule
  \end{tabular}
  \caption{Per-reviewer activity over the eight-reviewer campaign.
  \textit{Docs} and \textit{Pairs} count the documents and discrepant
  pairs each reviewer adjudicated, \textit{Anns} the annotations they
  saved, \textit{Flag rate} the share of their pairs flagged, and
  \textit{Changed vs ref.} the share of pairs whose decision differs
  from the reference. Rates are descriptive because reviewers were
  assigned different sets of conflict pairs. Reviewers are
  pseudonymous.}
  \label{tab:annotator-activity}
\end{table}

Across all annotation outcomes recorded by the campaign
(Table~\ref{tab:who-won}), reviewers accepted the reference version
unchanged in 12.7\% of outcomes and the model version unchanged in
0.6\%; field-level hybrids accounted for 63.2\% of the
outcomes and were by far the most common human-selected resolution
(82.7\% of the 3{,}150 outcomes resolved by a reviewer), while the
remaining 23.6\% are matched pairs on which both streams already
agreed and which were retained automatically.
The high hybrid rate confirms that the per-field adjudication path
is exercised in practice rather than being a rarely used affordance,
and that reviewers do not default to accepting the reference layer.

\begin{table}[!htbp]
  \centering
  \small
  \begin{tabular}{lrr}
  \toprule
  \textbf{Outcome} & \textbf{Anns} & \textbf{\%} \\
  \midrule
  Reference accepted unchanged & 522     & 12.7 \\
  Model accepted unchanged     & 24      & 0.6 \\
  Field-level hybrid   & 2{,}604 & 63.2 \\
  Both streams agreed  & 971     & 23.6 \\
  \midrule
  Total                & 4{,}121 & 100.0 \\
  \bottomrule
  \end{tabular}
  \caption{Distribution of final annotation outcomes recorded by the
  campaign. \textit{Both streams agreed} denotes automatically
  retained matched records that were not routed to reviewers.}
  \label{tab:who-won}
\end{table}

On the 107 annotations shared across the overlap subset, Fleiss'
$\kappa$~\citep{fleiss1971kappa} over adjudication decisions is
0.624 on all 78 shared pairs, and 0.720 on the 13 pairs that no
reviewer flagged (65 of the 78 overlap pairs drew at least one flag),
with per-field values that expose which parts of the schema are
hardest to reconcile (Table~\ref{tab:fleiss}).
On the full set, quantity and modifier show substantial agreement,
unit is fair, and event type is the ambiguous
frontier of the schema (Figure~\ref{fig:adjudication}), consistent
with the fact that most flagged
pairs push back on the same category boundary. Agreement is higher on
pairs that no reviewer flagged, indicating that items flagged by at
least one reviewer tend to be harder to adjudicate; the unflagged
result is descriptive, however, because it rests on only 13 pairs.
Reviewers themselves show essentially no agreement on \emph{which}
pairs warrant a flag (Fleiss' $\kappa = -0.01$ over a binary
flagged/unflagged label per pair). We therefore treat flagging as an
individual uncertainty signal and routing mechanism rather than as a
consensus annotation.
Agreement here measures adjudication consistency, not independent
annotation reliability.

\begin{table}[!htbp]
  \centering
  \small
  \begin{tabular}{lcc}
  \toprule
  \textbf{Field} & \textbf{$\kappa$ (all pairs)} & \textbf{$\kappa$ (unflagged)} \\
  \midrule
  Event type & 0.225 & 0.473 \\
  Quantity   & 0.688 & 0.717 \\
  Unit       & 0.337 & 0.529 \\
  Modifier   & 0.683 & 1.000 \\
  \midrule
  \textbf{Overall} & \textbf{0.624} & \textbf{0.720} \\
  \bottomrule
  \end{tabular}
  \caption{Fleiss' $\kappa$ over the five-reviewer overlap subset
  (10 shared documents), on all 78 shared pairs and on the 13 pairs no
  reviewer flagged. Agreement is higher on the unflagged pairs, a
  descriptive result given their small number; event type is the
  schema's hardest field.}
  \label{tab:fleiss}
\end{table}

The system is open source under the MIT license:
\url{https://github.com/mattemusacchio/quanreview}
(Python + Eel; launchers for macOS, Linux, and Windows; dependencies
pinned with uv; \texttt{run\_demo.sh} launches the tool on a bundled
sample dataset).

\section{Conclusions}

QuanReview turns the reconciliation of human and LLM annotation layers
from an ad-hoc scripting task into an auditable, offline workflow: it
aligns two streams at character level, resolves one-sided records by
explicit and logged policy, routes only genuine conflicts to a
browser-based adjudication interface, and exports a corrected layer in
the same JSON schema and file layout as the input, so it can replace
the original annotation files directly without changing any downstream
tooling. On a 4{,}457-record humanitarian
benchmark it auto-merged 8\% of documents, applied explicit policies
to a further 1{,}513 records, including 287 model-only records
deferred for re-annotation, and concentrated reviewers on 3{,}131
candidate conflicts, where field-level hybrids, rather than wholesale
acceptance of either stream, accounted for most decisions.

Two directions extend the system beyond its current scope. The first
is generalisability across annotation schemas: the comparison layer is
already schema-driven, and we are broadening it so that additional
schema-compatible typed-span tasks can be reconciled without touching
code. The second
is dynamic configuration, letting schemas, policies, and redundancy be
edited from the interface at run time rather than fixed in a static
file before a campaign begins. Together these would let QuanReview
serve as a general adjudication front-end for keeping annotation layers
trustworthy as models increasingly both produce and consume them.

\section{Limitations}

QuanReview adjudicates between two streams; it does not support
annotation from raw text, for which established platforms exist and
were used to create the reference layer in the first place. Agreement
is currently computed over adjudication decisions, which measures
reviewer consistency rather than independent annotation reliability.
The schema is configurable, but the system has so far been exercised
on one schema family, so we claim extensibility to other typed-span
tasks rather than demonstrated domain independence.

\section*{Ethics Statement}

Annotation campaigns can process sensitive humanitarian text. Because
the tool runs locally and stores only annotations and review
metadata, data governance stays with the dataset owner, and exported
logs identify reviewers by pseudonymous annotator IDs. The
adjudication interface labels the two streams neutrally
(\textit{Annotation 1} and \textit{Annotation 2}) and randomises their
left--right order independently for every pair, to limit anchoring on
the reference stream.

\bibliography{biblio}

@inproceedings{klie2018inception,
    title     = {The {INC}ep{TION} Platform: Machine-Assisted and Knowledge-Oriented Interactive Annotation},
    author    = {Klie, Jan-Christoph and Bugert, Michael and Boullosa, Beto and Eckart de Castilho, Richard and
  Gurevych, Iryna},
    booktitle = {Proceedings of the 27th International Conference on Computational Linguistics: System Demonstrations},
    year      = {2018},
    pages     = {5--9}
  }

@inproceedings{stenetorp2012brat,
    title     = {brat: a Web-based Tool for {NLP}-Assisted Text Annotation},
    author    = {Stenetorp, Pontus and Pyysalo, Sampo and Topi{\'c}, Goran and Ohta, Tomoko and Ananiadou, Sophia and
  Tsujii, Jun'ichi},
    booktitle = {Proceedings of the Demonstrations at the 13th Conference of the European Chapter of the Association
  for Computational Linguistics},
    year      = {2012},
    pages     = {102--107}
  }

@misc{nakayama2018doccano,
    title  = {doccano: Text Annotation Tool for Human},
    author = {Nakayama, Hiroki and Kubo, Takahiro and Kamura, Junya and Taniguchi, Yasufumi and Liang, Xu},
    year   = {2018},
    note   = {Software available from \url{https://github.com/doccano/doccano}}
  }

@misc{tkachenko2020labelstudio,
    title  = {Label {S}tudio: Data labeling software},
    author = {Tkachenko, Maxim and Malyuk, Mikhail and Holmanyuk, Andrey and Liubimov, Nikolai},
    year   = {2020--2024},
    note   = {\url{https://github.com/HumanSignal/label-studio}}
  }

@misc{montani2018prodigy,
    title  = {Prodigy: A modern annotation tool for creating training data for machine learning models},
    author = {Montani, Ines and Honnibal, Matthew},
    year   = {2018},
    note   = {\url{https://prodi.gy}}
  }

@misc{argilla2023,
    title  = {Argilla: Open-source data curation platform for {LLM}s},
    author = {{Argilla}},
    year   = {2023},
    note   = {\url{https://github.com/argilla-io/argilla}}
  }

@article{skitka1999automation,
    title   = {Does automation bias decision-making?},
    author  = {Skitka, Linda J. and Mosier, Kathleen L. and Burdick, Mark},
    journal = {International Journal of Human-Computer Studies},
    volume  = {51},
    number  = {5},
    pages   = {991--1006},
    year    = {1999}
  }

@article{fleiss1971kappa,
    author  = {Fleiss, Joseph L.},
    title   = {Measuring nominal scale agreement among many raters},
    journal = {Psychological Bulletin},
    volume  = {76},
    number  = {5},
    pages   = {378--382},
    year    = {1971},
    doi     = {10.1037/h0031619}
}

@inproceedings{fekih2022humset,
  title     = {{HumSet}: Dataset of Multilingual Information Extraction and Classification for Humanitarian Crises Response},
  author    = {Fekih, Selim and Tamagnone, Nicolo' and Minixhofer, Benjamin and Shrestha, Ranjan and Contla, Ximena and Oglethorpe, Ewan and Rekabsaz, Navid},
  booktitle = {Findings of the Association for Computational Linguistics: EMNLP 2022},
  pages     = {4379--4389},
  year      = {2022},
  address   = {Abu Dhabi, United Arab Emirates},
  publisher = {Association for Computational Linguistics},
  doi       = {10.18653/v1/2022.findings-emnlp.321}
}

@inproceedings{liberatore2024quantitative,
  title     = {Quantitative Information Extraction from Humanitarian Documents},
  author    = {Liberatore, Daniele and Kalimeri, Kyriaki and Sever, Derya and Mejova, Yelena},
  booktitle = {Proceedings of the 2024 International Conference on Information Technology for Social Good (GoodIT '24)},
  year      = {2024},
  address   = {Bremen, Germany},
  publisher = {ACM},
  doi       = {10.1145/3677525.3678667}
}

\end{document}